\documentclass{article}
\usepackage{spconf}
\usepackage{hyperref}

\usepackage{microtype}
\usepackage{inconsolata}
\usepackage{graphicx}
\usepackage{amsmath}
\usepackage{amssymb}
\usepackage{algorithm}
\usepackage{algorithmic}
\usepackage{mathtools}
\usepackage[numbers]{natbib}
\usepackage{xcolor} 
\usepackage{multicol}
\usepackage{multirow}
\usepackage{tabularx}
\usepackage{booktabs}
\usepackage{listings}
\usepackage{colortbl}
\definecolor{lightgreen}{RGB}{144, 215, 144}
\definecolor{lightgray}{RGB}{235, 235, 235}
\title{From Token to Line: Enhancing Code Generation with a Long-Term Perspective}
\name{
\parbox{\linewidth}{\centering
Tingwei Lu$^{1}$ \qquad Yangning Li$^{1,2}$ \qquad Liyuan Wang$^{3}$ \qquad Binghuai Lin$^{3}$ \\
Qingsong Lv$^{1}$ \qquad Zishan Xu$^{1}$ \qquad Hai-Tao Zheng$^{1,2,\dagger}$ \thanks{$\dagger$ Corresponding author: zheng.haitao@sz.tsinghua.edu.cn. This research is supported by National Natural Science Foundation of China (Grant No.62276154);
the Natural Science Foundation of Guangdong Province (Grant No.2024TQ08X729);
Basic Research Fund of Shenzhen City (Grant No.JCYJ20240813112009013 and GJHZ20240218113603006);
The Major Key Project of PCL for Experiments and Applications (Grant No.PCL2024A08).} \qquad Yinghui Li$^{3}$ \qquad Hong-Gee Kim$^{4}$}
}

\address{$^{1}$ Shenzhen International Graduate School, Tsinghua University \qquad  $^{2}$ Peng Cheng Laboratory \\
 $^{3}$ Tencent Technology Co., Ltd \qquad $^{4}$ Seoul National University \\
 \{ltw23, yn-li23\}@mails.tsinghua.edu.cn}

\begin{document}

\maketitle
\begin{abstract}
The emergence of large language models (LLMs) has significantly promoted the development of code generation task, sparking a surge in pertinent literature. Current research is hindered by redundant generation results and a tendency to overfit local patterns in the short term. Although existing studies attempt to alleviate the issue by adopting a multi-token prediction strategy, there remains limited focus on choosing the appropriate processing length for generations. By analyzing the attention between tokens during the generation process of LLMs, it can be observed that the high spikes of the attention scores typically appear at the end of lines. This insight suggests that it is reasonable to treat each line of code as a fundamental processing unit and generate them sequentially. Inspired by this, we propose the \textbf{LSR-MCTS} algorithm, which leverages MCTS to determine the code line-by-line and select the optimal path. Further, we integrate a self-refine mechanism at each node to enhance diversity and generate higher-quality programs through error correction. Extensive experiments and comprehensive analyses on three public coding benchmarks demonstrate that our method outperforms the state-of-the-art performance approaches.
\end{abstract}


\section{Introduction}
Large language models (LLMs) such as LLaMA~\cite{touvron2023llama} and GPT-4~\cite{achiam2023gpt} have achieved tremendous success across various domains recently, particularly in NLP~\cite{team2023gemini,qingsong2025raise, li2025ultrawiki,chen2025dast,li2025teaching}. The code generation task aims to automatically generate code meeting the requirements based on the provided natural language (NL) description, which can be regarded as a text sequence. Thus, code generation can still be considered a specialized form of text generation, with the emergence of LLMs tailored to coding, known as Code LLMs.

\begin{figure}[t]
  \includegraphics[width=\linewidth]{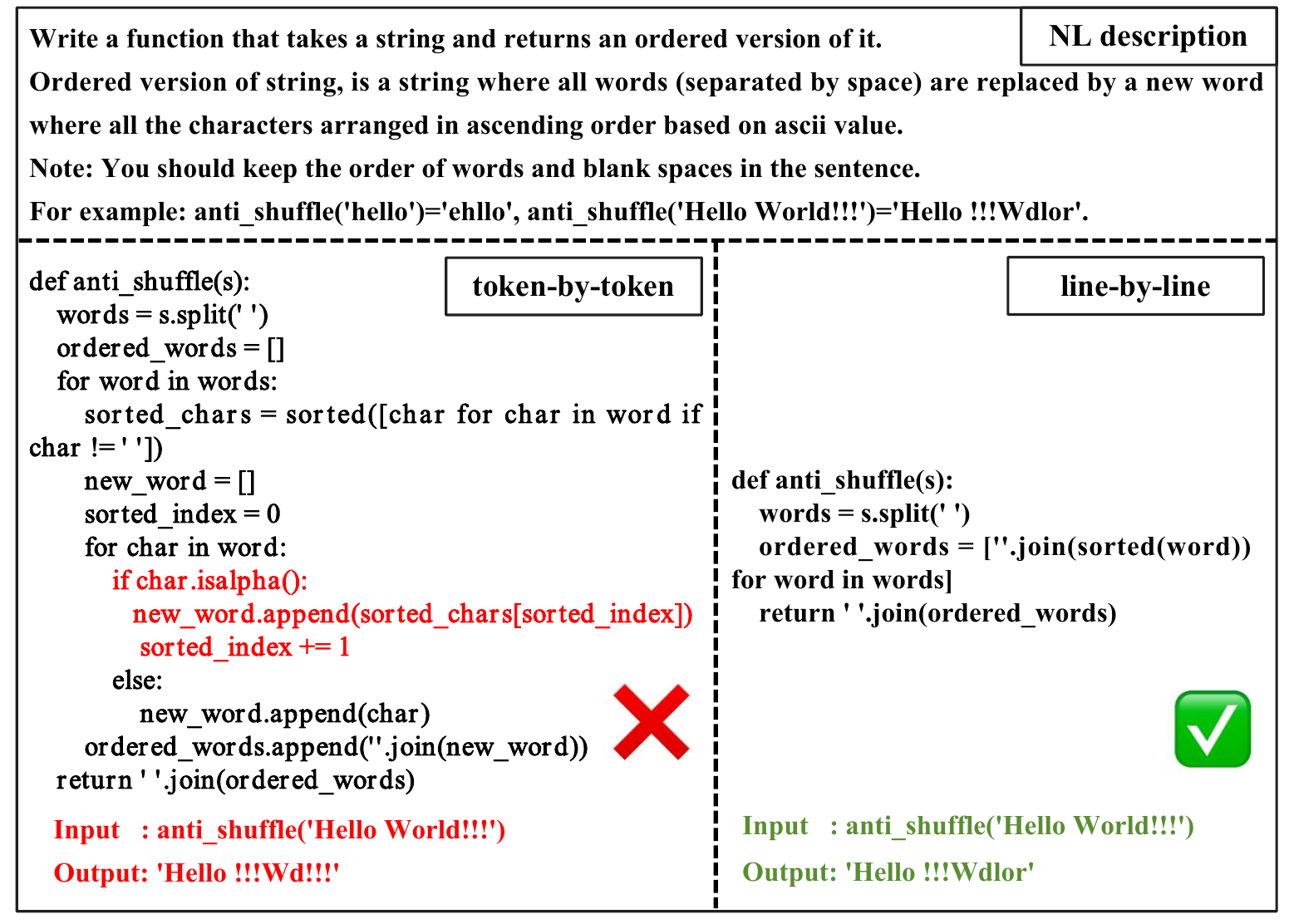}
  \vspace{-5mm}
  \caption {Examples of code generated by two kinds of methods.}
\vspace{-2mm}
\label{fig:case}
\end{figure}

\begin{figure}[ht]
  \includegraphics[width=\linewidth]{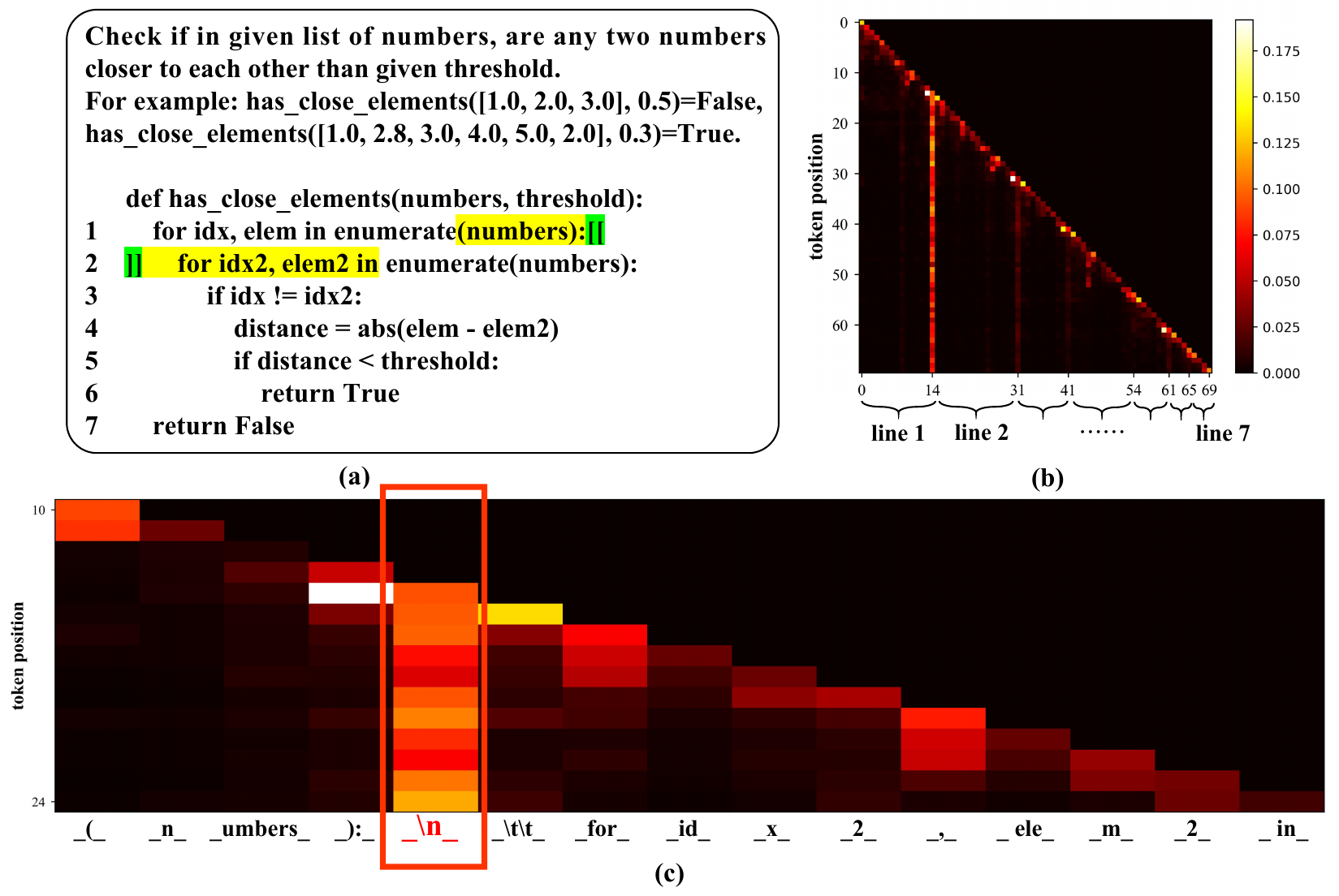}
  \vspace{-5mm}
  \caption {(a) A case is marked with line numbers on the left. (b) Global attention heatmap, where the range of each line is specifically annotated. The columnar appears at the end of each line. (c) Local attention maps for the yellow snippets of the code block, with each corresponding token labeled below the graph, and the line-end token `\textbackslash n' (in green) is particularly noticeable as a bar chart.}
\vspace{-2mm}
\label{fig:line_analyse} 
\end{figure}

The research on Code LLMs is divided into two prime avenues: (1) \textbf{Pre-train Code LLMs}. Pre-trained models such as CodeGen~\cite{nijkamp2022codegen}, StarCode~\cite{li2023starcoder}, and DeepSeek-Coder~\cite{guo2024deepseek} provide solid backbone for code tasks; (2) \textbf{Design decoding strategy}. Numerous decoding strategies~\cite{zhang2023planning,zhu2024hot} are proposed to correct errors generated by greedy decoding during inference. These methods are promising for their plug-and-play manner. We focus on the second in this paper.


Existing methods primarily generate code token-by-token using LLMs~\cite{zhang2023planning,brandfonbrener2024vermcts}, which pay more attention to short-term tokens at each generation. However, due to the strict logical structure and closely related knowledge inherent in programming languages, overlooking the long-term dependency on code may lead to severely flawed programs. Therefore, the token-level approaches, which concentrate on local code segments, are prone to misaligning code fragments with the natural language (NL) description or producing redundancy among the long-term perspective, as depicted in Figure \ref{fig:case}.

To overcome the short-term issue, Gloeckle et al.~\cite{gloeckle2024better} explore introducing multi-tokens prediction as an auxiliary training task, which encourages LLMs to consider longer-term dependencies within the generated sequence. The paper highlights the significance of attention between distant tokens for LLMs.
Following this work, we take a deeper dive into the attention between tokens of existing LLMs. As shown in Figure \ref{fig:line_analyse}, it is observed that certain tokens have a profound influence on subsequent generations. It can be inferred that these tokens can summarize information from the prior code and lead the following generation, which is denoted as \textbf{``summary tokens''}. Thus, ensuring the correctness of the previous summary token and the corresponding line is crucial, which is beneficial for future generations and rectification. The observation highlights that \textbf{the line emerges as a more effective fundamental processing unit in the code generation task}.

Motivated by this, we introduce a novel decoding strategy \textbf{L}ine-level \textbf{S}elf-\textbf{R}efine \textbf{M}onte \textbf{C}arlo \textbf{T}ree \textbf{S}earch, termed \textbf{LSR-MCTS}. It combines the line-level concept with MCTS, where each node in the tree signifies a line segment. A trajectory from the root to the leaf forms a complete program. LSR-MCTS shortens the distance between tokens in the tree from a higher horizon and encourages the model to predict from a global optimization, generating more concise programs depicted in Figure \ref{fig:case}. However, the higher horizon may overlook some viable branches. Considering that summary tokens can facilitate code correction, we integrate a self-refine mechanism at each node to regenerate the current line and summary token. Extensive experiments and comprehensive analysis validate the exceptional performance of the LSR-MCTS decoding strategy.


\section{Related Work}
\subsection{LLMs for Code}
LLMs have demonstrated remarkable capabilities in handling tasks such as NLP~\cite{tang2025gmsa, tang2025perception,anil2023palm,jiang2023mistral,jiang2024mixtral}. Numerous studies have shown that excessively long input text leads to performance degradation~\cite{tang2026read, tang2026comi, lv2026data, zhao2026comet, zhao2025cos}. As for the field of code generation, models like Codex~\cite{chen2021evaluating}, trained across a multitude of programming languages and billions of lines of code, have emerged as versatile code snippet generators, integrated into tools like Copilot to assist programmers in coding. AlphaCode~\cite{li2022competition}, which is trained on a vast array of open-source Python code, stands out as the first LLM capable of generating structured code directly from NL descriptions.

Stimulated by these pioneering efforts, many researchers are dedicated to the training of Code LLMs. Google introduces the proprietary PaLM-Coder~\cite{chowdhery2023palm}, which generates code results through API calls, showcasing impressive performance. Concurrently, other researchers focus on developing open-source Code LLMs, such as Salesforce's CodeGen~\cite{nijkamp2022codegen}, Meta's In-Coder~\cite{fried2022incoder}, Code Llama~\cite{roziere2023code}, and others including StarCoder~\cite{li2023starcoder}, CodeGeeX~\cite{zheng2023codegeex}, DeepSeek-Coder~\cite{guo2024deepseek}, etc. These models are progressively approaching and surpassing the performance of general models, bolstering the confidence in training Code LLMs and amplifying code generation efficiency. 

\begin{figure*}[th]
  \centering
  \vspace{-5mm}
  \includegraphics[width=0.9\linewidth]{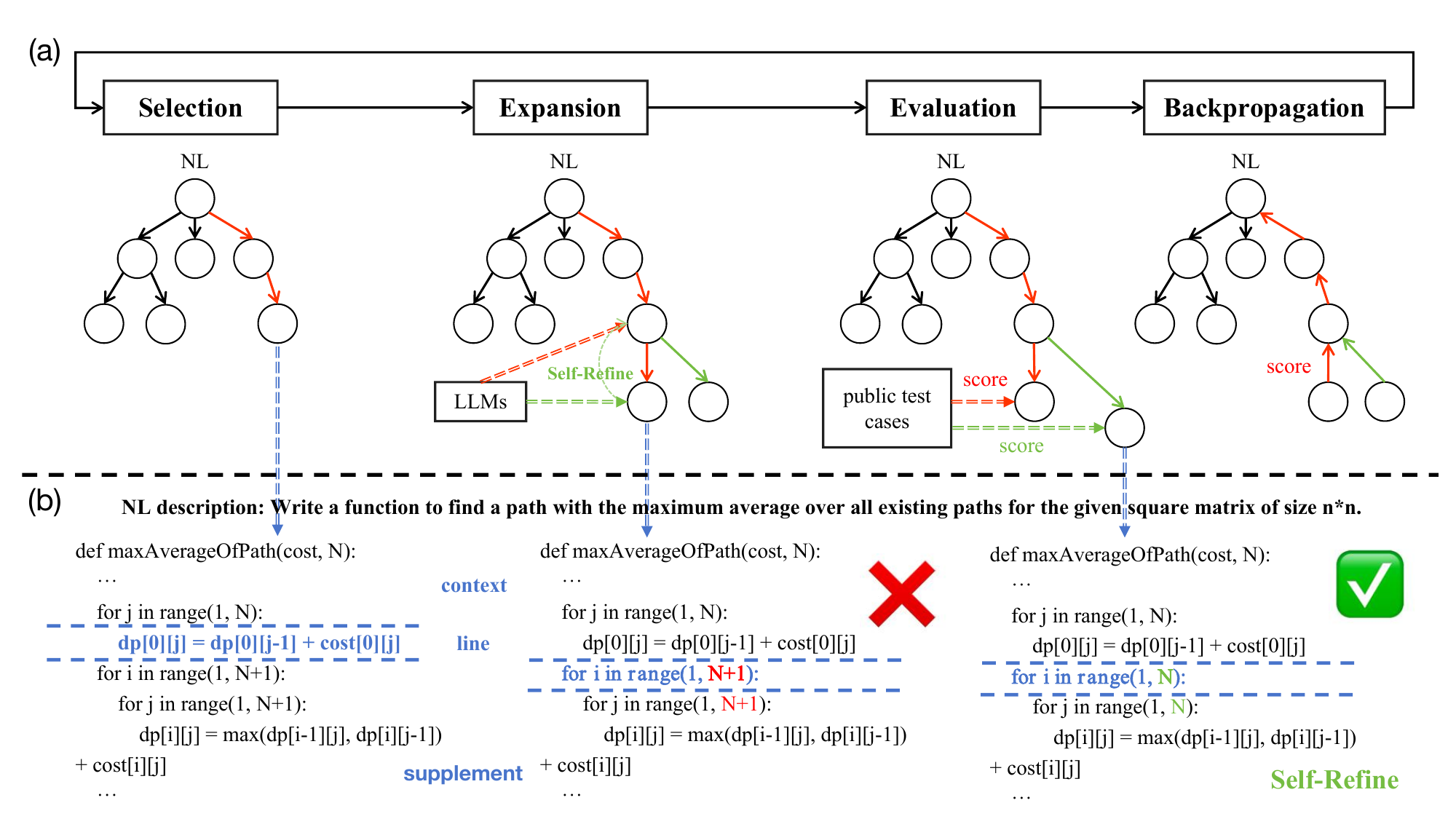}
  \caption {The framework of LSR-MCTS. The red part in (a) shows the four iterative steps of LSR-MCTS: selection, expansion, evaluation, and backpropagation. The green sections reflect the self-refine process, where new nodes are generated in the expansion step. Part (b) explicitly displays the content of a single node, including context, line, and supplement, with the main body "line" emphasized in bold blue.}
\vspace{-2mm}
\label{fig:lsr-mcts}
\end{figure*}

\subsection{Monte Carlo Tree Search}
The performance improvement of LLMs on various tasks is attributable to not merely their augmented capabilities from training, but also the optimization of their generation strategies~\cite{yasunaga2023large,chuang2023dola,huang2024opera}. MCTS, as one of the efficient strategies for handling large-scale search spaces, is highly applicable in the generation domain and is becoming a research hotspot in code generation.

 VerMCTS~\cite{brandfonbrener2024vermcts} designs a logical verifier within the MCTS process, expanding tokens until the verifier can return a score. Furthermore, PG-TD~\cite{zhang2023planning} proposes an MCTS-based method evaluated by test cases for code generation, treating each token decoded by LLMs as an action. However, the application of MCTS in code generation is predominantly token-level, focusing on short-term predictions, which causes a local optimal solution, especially when dealing with programs that consist of thousands of tokens. 
 As the distance between nodes increases with the length of the code, the practicality diminishes considerably.

\subsection{Self-Refine Strategy}
In addition to MTCS, self-refine can also be considered an efficient strategy~\cite{yao2023react,madaan2023self}. LATS~\cite{zhou2023language} employs a generation strategy that combines MCTS with self-reflection and environmental feedback, generating multiple programs from the same node and using prompts to reflect on incorrect code predictions. Reflexion~\cite{shinn2024reflexion} continuously refines and regenerates the code based on the environment—the textual feedback from LLMs on the generated code—ceasing until the evaluation metrics reach a plateau.

Nevertheless, the existing approaches are focused on generating introspection in the form of text or reconstructing entirely new code segments. These methodologies lack the ability to target and rectify localized errors accurately. Consequently, by employing a self-refine strategy at each node in the line-level MCTS, it is possible to pay close attention to local details while taking a global perspective.


\section{Methodology}
In this section, we elaborate on the proposed training-free decoding strategy LSR-MCTS. The comprehensive depiction of the entire process is shown in Figure \ref{fig:lsr-mcts}.


\subsection{Line-level MCTS}
MCTS treats code generation task as a meticulous process of tree search. The root node lies the initial Natural Language prompt describing the problem, with each subsequent node representing an extension of the generated code. The search space encompasses all conceivable branches of the tree. The objective of MCTS is to navigate this potentially unbounded search space by identifying the optimal child nodes, gradually constructing a coherent path that culminates in the complete and correct code solution.

LSR-MCTS framework adheres to the conventional MCTS algorithm's four-phase structure—selection, expansion, evaluation, and backpropagation. It enriches each phase with a line-level concept, enhancing the precision of the search. We redefine the node information within the tree structure to suit our line-level decoding process. As illustrated in Figure \ref{fig:lsr-mcts}(b), a node encompasses \textbf{context, line, and supplement}, which together form a segment of complete and executable code. The line represents a specific line of code that characterizes the node, while the context is an n-line fixed code block constructed from the path of ancestor nodes. To ensure that each node can be evaluated using public test cases, incomplete code blocks must be supplemented. Here, both the line and supplement are generated by LLMs, with the next line of the current node being selected as the line, and the rest as the supplement.

During the \textbf{selection phase}, we apply the upper confidence bound for trees(UCT) strategy, starting from the root node to the max score leaf node $s$:
\begin{align}
\operatorname{UCT(s)}=\frac{s.values}{s.visits}+c \cdot \sqrt{\frac{\ln N}{s.visits}}
\label{equal:uct}
\end{align}
where $s.values$ is the cumulative score of node $s$, $s.visits$ is the count of times node $s$ has been visited, and $N$ represents the number of rollouts that have been executed.
In the subsequent \textbf{expansion phase}, LLM is utilized to generate $m$ code block for leaf in non-terminating states, segmenting the complete code block $C$ into context, line, and supplement. Here $m$ is set to 3 for constraining the number of child nodes.
Upon acquiring the details of the next node, it is appended as a child to the current node. The quality of the generated code is then \textbf{evaluated} by the pass rate of public test cases.
Once the score of the code is determined, it is \textbf{backpropagated} to the root node, updating the value and the visited count of each ancestor node along the path, accordingly promoting future decision-making.
To improve the efficiency of code generation, we add a cache mechanism to store the code blocks of ancestor nodes.

\begin{algorithm}[t] 
\small
\caption{Line-level Self-Refine MCTS}
\label{algorithm}
\begin{algorithmic}[1] 
\REQUIRE ~~\\
$M_\theta$: LLM with parameters $\theta$; $root$: the root node of the tree; $m$: the maximum number of children of any node; $n$: the number of max rollouts; $PR$: the general generation prompt; $SRP$: self-refine PROMPT; $R(\cdot)$: reward function for code according to the public test cases.

\FOR{$i \gets 1$ \TO $n$}
    \STATE $node \gets root$; 
    
    \STATE \color{lightgreen}\# Selection \color{black} \label{selection:start}
    \WHILE{$|node.children|>0$}
        \STATE $node \gets UCT(node.children)$; \label{uct_function}
    \ENDWHILE \label{selection:end}
    
    \STATE \color{lightgreen}\# Expansion \color{black} \label{expansion:start}
    \STATE $next\_codes \gets M_\theta(PR+node.context, m)$;
    \FOR{$next\_code \in next\_codes$}
        \STATE $next\_node \gets$ Concat$(node, next\_code)$;
        \STATE Add $next\_node$ to the children of $node$;
        \STATE
        \STATE $refined\_code \gets M_\theta(SRP+next\_node, 1)$; \label{refine:start}
        \STATE $refined\_node \gets$ Concat$(node, refined\_code)$;
        \STATE Add $refined\_node$ to the $node.children$; \label{refine:end}
    \ENDFOR
    \label{expansion:end}

    \STATE \color{lightgreen}\# Evaluation \color{black} \label{evaluation:start}
    \STATE $r\_next \gets R(next\_node)$;
    \STATE $r\_refine \gets R(refined\_node)$; \label{evaluation:end}
    
    \STATE \color{lightgreen}\# Backpropagation \color{black} \label{backpropagation:start}
    \WHILE{$node.parent$}
        \STATE $node.values+=r\_next+r\_refine$
        \STATE $node.visits+=2$
        \STATE $node \gets node.parent$
    \ENDWHILE ~\label{backpropagation:end}
\ENDFOR
\STATE \color{lightgreen}\# Return \color{black}
\STATE $node \gets root$; 
\WHILE{$|node.children|>0$}
    \STATE $node \gets UCT(node.children)$;
\ENDWHILE
\RETURN $node$
\end{algorithmic}
\end{algorithm}

\subsection{Self-Refine Mechanism}
\label{sec:self-refine}
To address omissions of feasible branches due to the limitation on the number of child nodes, and to rectify the code to guarantee the precision of summary tokens that exert substantial influence on later generations, we introduce a self-refine mechanism during the expansion phase of Line-level MCTS.


This mechanism generates a new code block for any node whose score is low and for the last node in the current path. The newly created block acts as an unconstrained child of the current node, enabling further expansion in subsequent operations. The score threshold is empirically set to 0.5 to balance exploration and efficiency.
Once the new refined nodes are obtained, they are processed in the same manner as regular nodes during the evaluation and backpropagation stages, ultimately generating a superior program. It can be seen in Figure \ref{fig:lsr-mcts} (b) that the first two codes are incorrect. Through self-refine, they can be adjusted to the correct code.

\section{Experiments}

\begin{table*}[thb]
    \small
    \centering
    \setlength{\tabcolsep}{4 mm}
    \setlength{\extrarowheight}{1 pt}
    \resizebox{1\textwidth}{!}{
    \begin{tabular}{l|ccc|ccc|ccc}
    \toprule
        \multirow{2}{*}{Methods} &  \multicolumn{3}{@{}c}{{\bf HumanEval}} &  \multicolumn{3}{@{}c}{{\bf MBPP}} & \multicolumn{3}{@{}c}{{\bf Code Contests}} \\
        & pass@1 & pass@3 & pass@5 & pass@1 & pass@3 & pass@5 & pass@1 & pass@3 & pass@5 \\
    \midrule
    \midrule  
        \multicolumn{10}{@{}c}{{ \textit{Code-Specific Models}}} \\
    \midrule
\multicolumn{10}{@{}l}{{ \bf CodeLlama-7B-Instruct}} \\
Beam-Search & 36.1 & 39.4 & 40.2 & 30.1 & 32.9 & 33.6 & 6.7 & 8.8 & 9.5 \\
Top-p & 36.5 & 38.7 & 39.9 & 30.5 & 33.2 & 34.3 & 7.2 & 8.9 & 9.4 \\
Reflexion & 40.2 & 42.1 & 44.3 & 33.6 & 35.1 & 37.2 & 7.2 & 9.6 & 10.3 \\
PG-TD(T-MCTS) & 42.2 & 46.3 & 47.9 & 36.6 & 38.3 & 40.7 & 8.9 & 10.0 & 11.1 \\
TSR-MCTS & 43.5 & 47.4 & 48.2 & 37.8 & 39.0 & 41.6 & 9.3 & 10.7 & 11.5 \\
L-MCTS & 43.8 & 48.1 & 48.3 & 38.5 & 40.7 & 42.5 & 9.2 & 10.4 & 11.1 \\
\rowcolor{blue!10} LSR-MCTS & 45.7 & 49.4 & 50.6 & 40.8 & 42.2 & 43.9 & 9.8 & 11.2 & 12.0 \\
\cmidrule (r){1-1} \cmidrule (lr){2-4} \cmidrule (lr){5-7} \cmidrule (lr){8-10}

\multicolumn{10}{@{}l}{{ \bf aiXcoder-7B}} \\
Beam-Search & 47.0 & 51.7 & 52.9 & 40.9 & 46.3 & 48.0 & 8.2 & 9.4 & 10.2 \\
Top-p & 47.3 & 51.9 & 53.4 & 40.3 & 46.0 & 47.5 & 8.2 & 9.6 & 10.6 \\
Reflexion & 48.6 & 52.3 & 54.5 & 44.8 & 48.0 & 49.3 & 9.6 & 10.7 & 11.4 \\
PG-TD & 50.1 & 54.6 & 55.9 & 46.1 & 49.4 & 50.2 & 10.1 & 11.3 & 12.1 \\
\rowcolor{blue!10} LSR-MCTS & 53.3 & 57.8 & 58.1 & 48.3 & 51.7 & 53.3 & 11.6 & 12.8 & 13.6 \\
    \midrule
    \midrule
        \multicolumn{10}{@{}c}{{ \textit{General Models}}} \\
    \midrule
\multicolumn{10}{@{}l}{{ \bf GPT-4}} \\ 
Beam-Search & 85.4 & 86.7 & 87.2 & 49.1 & 49.9 & 50.2 & 12.3 & 14.3 & 15.2 \\
Top-p & 86.5 & 87.2 & 87.7 & 49.8 & 50.6 & 51.1 & 12.5 & 14.3 & 15.5 \\
Reflexion & 88.6 & 90.4 & 91.1 & 51.6 & 52.3 & 53.4 & 13.7 & 15.2 & 16.5 \\
PG-TD(T-MCTS) & 89.3 & 90.7 & 91.3 & 53.2 & 54.4 & 55.8 & 14.7 & 15.4 & 16.3 \\
TSR-MCTS & 89.7 & 90.9 & 91.5 & 53.6 & 54.8 & 55.9 & 14.9 & 15.9 & 16.8 \\
L-MCTS & 89.7 & 91.4 & 92.4 & 53.7 & 54.6 & 56.3 & 15.2 & 16.0 & 16.8 \\
\rowcolor{blue!10} LSR-MCTS & 90.6 & 92.3 & 93.1 & 54.9 & 55.8 & 57.1 & 16.2 & 17.3 & 17.9 \\
\cmidrule (r){1-1} \cmidrule (lr){2-4} \cmidrule (lr){5-7} \cmidrule (lr){8-10}

\multicolumn{10}{@{}l}{{ \bf Llama3-8B-Instruct}} \\
Beam-Search & 62.0 & 64.1 & 64.7 & 30.1 & 32.9 & 33.8 & 6.6 & 8.0 & 9.2 \\
Top-p & 62.7 & 65.2 & 65.6 & 29.7 & 32.3 & 33.1 & 7.1 & 7.9 & 8.9 \\
Reflexion & 66.7 & 68.2 & 70.4 & 34.6 & 36.7 & 37.7 & 7.3 & 8.8 & 9.5 \\
PG-TD & 67.3 & 69.6 & 71.5 & 36.6 & 38.3 & 39.1 & 8.2 & 9.7 & 10.0 \\
\rowcolor{blue!10} LSR-MCTS & 70.2 & 73.3 & 73.9 & 38.2 & 39.7 & 42.2 & 9.6 & 10.3 & 11.1 \\
    \bottomrule
    \end{tabular}
    }
    \vspace{-2mm}
    \caption{Main results and ablation performance on three public benchmarks. The best results are highlighted in light blue.}
\vspace{-2mm}
\label{tab:main}
\end{table*}

\subsection{Experimental Setup}
\noindent\textbf{Dataset.} Three commonly public Python code datasets are chosen for analysis, including \textbf{HumanEval}\footnote{\url{https://huggingface.co/datasets/openai/openai\_humaneval}}~\cite{chen2021evaluating} and \textbf{MBPP}\footnote{\url{https://huggingface.co/datasets/google-research-datasets/mbpp}}~\cite{austin2021program} of foundational difficulty and \textbf{Code Contests}\footnote{\url{https://huggingface.co/datasets/deepmind/code\_contests}}~\cite{li2022competition} of competitive programming difficulty. Each dataset comprises a natural language description of a programming problem, associated test cases and manually crafted solutions.


\noindent\textbf{Models.}
Two categories of LLMs are utilized to evaluate our proposed method. The first category is public code-specific LLMs, including \textbf{CodeLlama-7B-Instruct}~\cite{roziere2023code} and \textbf{aiXcoder-7B}. To demonstrate the generalizability of LSR-MCTS, extensive experiments are also conducted on general LLMs \textbf{GPT-4}~\cite{achiam2023gpt} and \textbf{Llama3-8B-Instruct}~\cite{dubey2024llama}.

\noindent\textbf{Baselines}
are categorized into three groups: traditional decoding methods, self-refine methods, and MCTS-based methods. \textbf{Beam-search} and \textbf{Top-p} are chosen as the traditional baselines, which are widely used in generation tasks. For the self-refine method, \textbf{Reflexion}~\cite{shinn2024reflexion} is adopted. It continuously refines and regenerates the code based on the textual feedback from LLMs until convergence. For the MCTS-based method, \textbf{PG-TD}~\cite{zhang2023planning} is selected, which is a token-level MCTS approach. The hyperparameter c in Equation \ref{equal:uct} is set to 4, and the rollout n is set to 100 for comparison.

\noindent\textbf{Metric.}
$pass@k$~\cite{chen2021evaluating} is used to assess the functional correctness of code generated by LLMs, where $k$ code samples are produced for each problem, with $k=1,3,5$. We generate $n \geq k$ programs for each data, and $c$ of them pass the private test cases. The unbiased estimate is calculated:
\begin{equation*}
pass@k:=\underset{Problems}{\mathbb{E}}\left[1-\frac{\binom{n-c}{k}}{\binom{n}{k}}\right]
\label{equal:passk}
\end{equation*}

\subsection{Main Experiments}

The main experimental results presented in Table \ref{tab:main} highlight the significant performance advantages of LSR-MCTS across various evaluation metrics and benchmarks, demonstrating its robust generalization capabilities. LSR-MCTS excels in all aspects, better handling a multitude of coding tasks.

Compared to code-specific models, Llama3 shows a higher level of performance on the HumanEval dataset, but not on Code Contests. This advantage can be attributed to the multi-task training scheme adopted by general models, giving them an enhanced ability to comprehend simple natural language problem descriptions. However, as the difficulty increases, they are hard to capture the close connections between code tokens, in which case Code LLMs are more applicable.

A more in-depth analysis of the dataset reveals that these models demonstrate greater enhancement on the challenging competitive programming dataset Code Contests, as opposed to the normal difficulty of HumanEval and MBPP. Particularly noteworthy is the significant improvement of 12.4\% for aiXcoder at pass@5, indicating that LSR-MCTS is adept at stimulating the potential of LLMs on complex issues.

As the value of $k$ changes, the proportion of enhancement by LSR-MCTS for the same model and dataset remains generally stable. The unbiased estimation characteristic of $pass@k$ suggests that the model exhibits excellent robustness.

\subsection{Ablation Study}

We designed experiments to analyze the influence of the two integral components of LSR-MCTS on model performance. Table \ref{tab:main} shows a comparative analysis on CodeLlama and GPT-4. \textbf{T-MCTS} eliminates the line-level strategy and only considers token-level MCTS, i.e., PG-TD. \textbf{TSR-MCTS} adds a self-refine mechanism on the basis of token-level. In contrast, \textbf{L-MCTS} removes the self-refine module from LSR-MCTS. Here, the MCTS rollout is set to 100 for all.

The comparison results reveal that, under identical hyperparameter settings, both components are instrumental in enhancing the performance of the decoding strategy. Interestingly, the contribution of the self-refine module at the token-level is negligible, with almost no discernible improvement. This stands in sharp contrast to the significant enhancement observed at the line-level, probably because the token-level approach may diminish the semantic connections between tokens over long distances, which is precisely what self-refine needs to exploit. Therefore, the combination of the two does not yield benefits. This observation emphasizes the synergy between line-level MCTS and the self-refine mechanism, which can reinforce each other effectively.


\subsection{Parameter Analysis}
\begin{figure}[t]
  \includegraphics[width=0.98\linewidth]{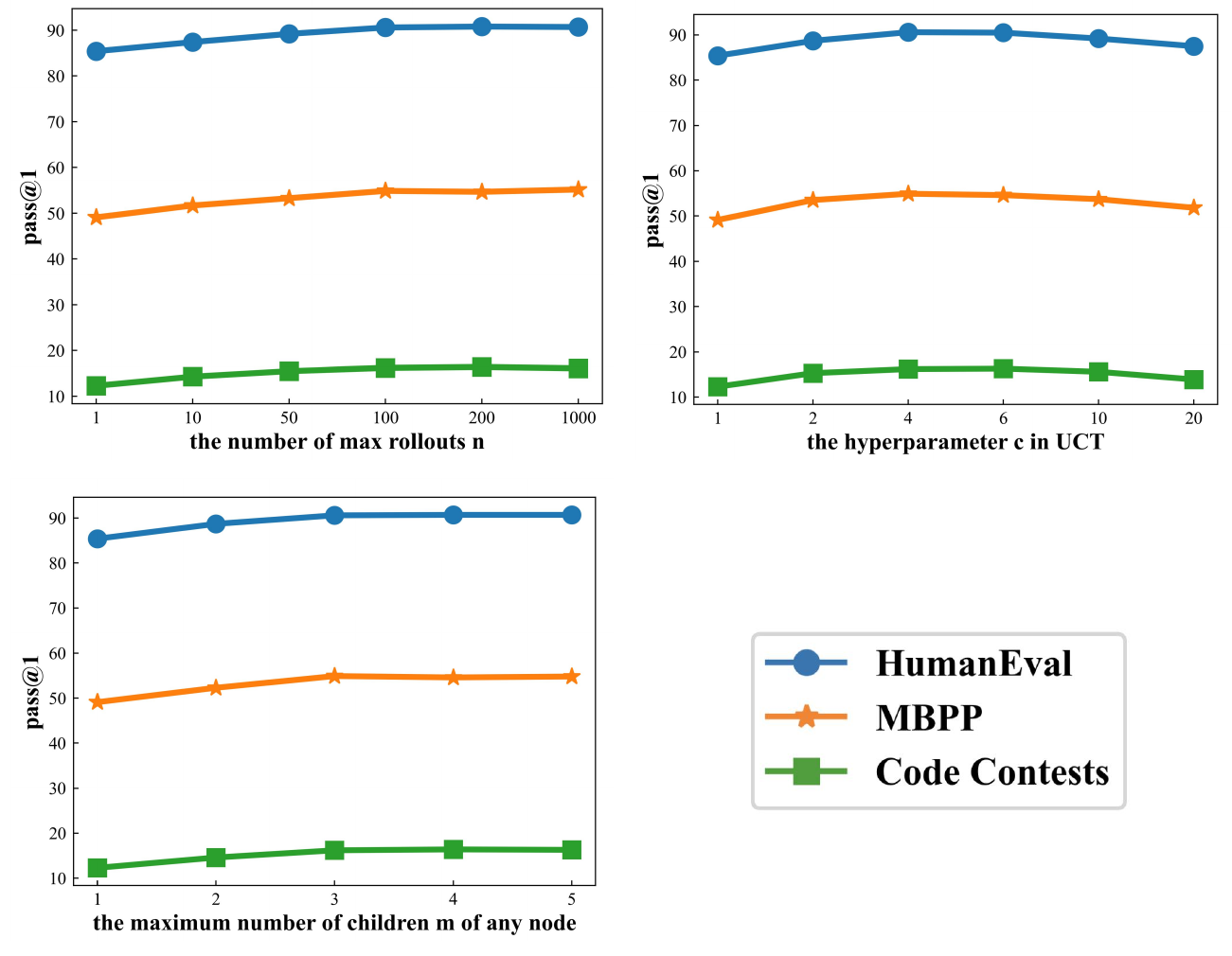}
  \caption {The impact of hyperparameter variations on GPT-4 performance. Hyperparameters include the number of max rollouts $n$, the UCT parameter $c$, and the maximum number of child nodes $m$ in the tree.}
\label{fig:parameters} 
\end{figure}

To investigate the sensitivity of LSR-MCTS to its hyperparameters, extensive experiments are conducted by varying the maximum rollouts $n$, parameter $c$ in UCT, and max children node count $m$. Based on the information depicted in Figure \ref{fig:parameters}, the following can be analyzed:

(1) The parameter $n$ governs the number of expansion iterations during the simulation process of the model. When $n=1$, no search is conducted, and the program is generated directly, which is akin to beam search. As $n$ increases, there is a noticeable enhancement in model performance, which eventually plateaus. This is because, in the initial phase, the model rapidly improves performance by increasing the number of simulations. However, after reaching a certain threshold, the potentiality of the model is fully activated, and no further improvements are observed.

(2) In the UCT algorithm, the parameter $c$ is utilized to balance the exploration and exploitation within the search process. A higher value of $c$ encourages the model to delve into nodes that have not been thoroughly explored, which may lead to excessive exploration and a consequent decline in performance. Conversely, a lower value of $c$ inclines the model to capitalize on known optimal paths, potentially causing the model to converge prematurely and miss out on optimal solutions. Thus, a moderate $c$ value can enable the model to achieve peak performance.

(3) The hyperparameter $m$ influences the search space of the model by limiting the number of child nodes. MCTS follows a single path when $m=1$, which is essentially beam search. Incrementing $m$ allows the model to explore a greater number of nodes at each junction, potentially enhancing the accuracy of code generation. Nevertheless, this also escalates the computational complexity. As shown in Figure \ref{fig:parameters}, model performance initially improves with the increase of $m$ and then stabilizes, indicating that augmenting $m$ within a certain limit can boost the capability of the model. However, beyond a certain value, additional nodes do not significantly enhance performance.

\section{Conclusion}
In this paper, we present a novel training-free decoding strategy called LSR-MCTS. This strategy consists of a line-level MCTS and a self-refine mechanism. The former segments the code block of each node into context, line, and supplement, generating the code line-by-line from a global perspective. The self-refine mechanism is employed to discover more effective programs and rectify code blocks. Extensive experiments conducted on three benchmarks demonstrate that LSR-MCTS achieves state-of-the-art performance across all models.

\bibliographystyle{IEEEbib}
\bibliography{refs}

\end{document}